\def\year{2021}\relax
\documentclass[letterpaper]{article} 
\usepackage{aaai21}  
\usepackage{times}  
\usepackage{helvet} 
\usepackage{courier}  
\usepackage[hyphens]{url}  
\usepackage{graphicx} 
\usepackage{natbib}  
\usepackage{caption} 
\usepackage{multirow}
\usepackage{amsmath}
\usepackage{amssymb}
\usepackage{arydshln}
\usepackage{xcolor}
\usepackage{multirow}
\usepackage{multicol}
\usepackage{graphicx}
\usepackage{subfigure} 
\usepackage{amsmath} 

\title{Interpretable NLG for Task-oriented Dialogue Systems with \\ Heterogeneous Rendering Machines}
\author{
	Yangming Li\textsuperscript{\rm 1,2}\footnote{This work was done when the first author did internship at Ant Group.},
	Kaisheng Yao\textsuperscript{\rm 2} \\
}
\affiliations{
	\textsuperscript{\rm 1}Harbin Institute of Technology \\
	\textsuperscript{\rm 2}Ant Group \\
	yangmingli@ir.hit.edu.cn, kaisheng.yao@antgroup.com
}

\begin{document}
	\maketitle
	
\begin{abstract}
		
	End-to-end neural networks have achieved promising performances in natural language generation (NLG). However, they are treated as black boxes and lack interpretability. To address this problem, we propose a novel framework, heterogeneous rendering machines (HRM), that interprets how neural generators render an input dialogue act (DA) into an utterance. HRM consists of a renderer set and a mode switcher. The renderer set contains multiple decoders that vary in both structure and functionality. For every generation step, the mode switcher selects an appropriate decoder from the renderer set to generate an item (a word or a phrase). To verify the effectiveness of our method, we have conducted extensive experiments on 5 benchmark datasets. In terms of automatic metrics (e.g., BLEU), our model is competitive with the current state-of-the-art method. The qualitative analysis shows that our model can interpret the rendering process of neural generators well. Human evaluation also confirms the interpretability of our proposed approach.
		
\end{abstract}
	
\section{Introduction}
	
\begin{figure}[t]
		\centering
		\includegraphics[width=0.45\textwidth]{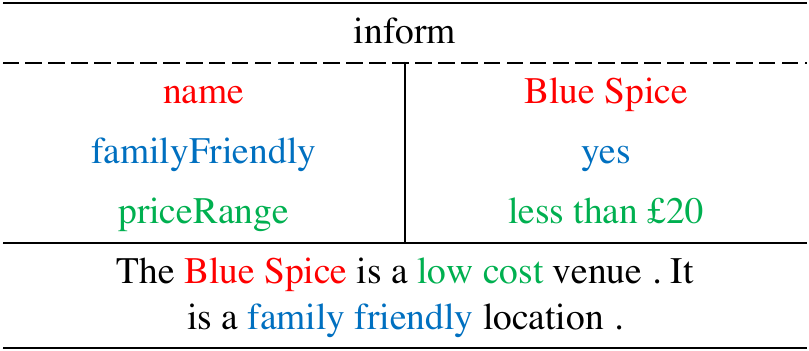}
		
		\caption{This example case is extracted from E2E-NLG dataset~\citep{novikova-etal-2017-e2e}. The upper part is a DA and the lower part is an utterance.}
		\label{fig:Data Case}
	\end{figure}

	Natural language generation (NLG), a critical component of task-oriented dialogue systems, converts a meaning representation, i.e., dialogue act (DA), into a natural language utterance. As demonstrated in Figure \ref{fig:Data Case}, an input DA consists of an act type and a set of slot values, while the ground truth is a sequence of words. We categorize all the slots into three types: 1) delexicalizable slot whose value is always propagated verbatim to the utterance (colored in red); 2) indicative slot whose value is yes, no, and so on (blue); 3) reworded slot whose value is possibly paraphrased into another expression in the utterance (green). Note that the slot types are defined by us for better understanding and this information is not available in the datasets.
	
	Conventional approaches~\citep{mirkovic2011dialogue,cheyer2014method} are mostly pipeline-based, dividing NLG tasks into sentence planning and surface realization. Sentence planning determines a tree-like structure of the given input DA, while surface realization linearizes the structure into the final surface form. Although these models are of great interpretability, they heavily rely on handcraft rules and domain-specific knowledge. Recently, data-driven methods using end-to-end neural networks have attracted much attention~\citep{tran-nguyen-2017-natural,zhu2019multi,li-etal-2020-slot}. For example, \citet{dusek-jurcicek-2016-sequence} apply sequence-to-sequence learning~\citep{bahdanau2014neural} to model response generation. These methods facilitate end-to-end learning on the unaligned corpus and achieve state-of-the-art results. However, they are treated as black boxes and lack interpretability. In this work, we consider a neural generator is interpretable if it's capable of showing how slot values are rendered and their locations in the generated utterance.
	
	To generalize NLG models to low-frequency and unseen slot values, prior methods mostly adopt the delexicalization technique, where slot values in the utterance are replaced with corresponding placeholders~\citep{wen-etal-2015-stochastic,wen-etal-2015-semantically,tran-nguyen-2017-natural}. While delexicalization, to some extent, has improved the interpretability, it's applicable for delexicalizable slots only, and therefore has limited usage. This is especially apparent in datasets with the majority of slots being reworded, e.g., E2E-NLG dataset~\citep{novikova-etal-2017-e2e}. Moreover, \citet{nayak2017plan,juraska-etal-2018-deep} observe that using delexicalization can result in lexical errors. Visualizing attention weight is another approach to understanding neural generators. However, many works \citep{jain-wallace-2019-attention} have found that it's hard to reach meaningful interpretability.
	
	\begin{figure*}[h]
		\centering
		\includegraphics[width=0.8\textwidth]{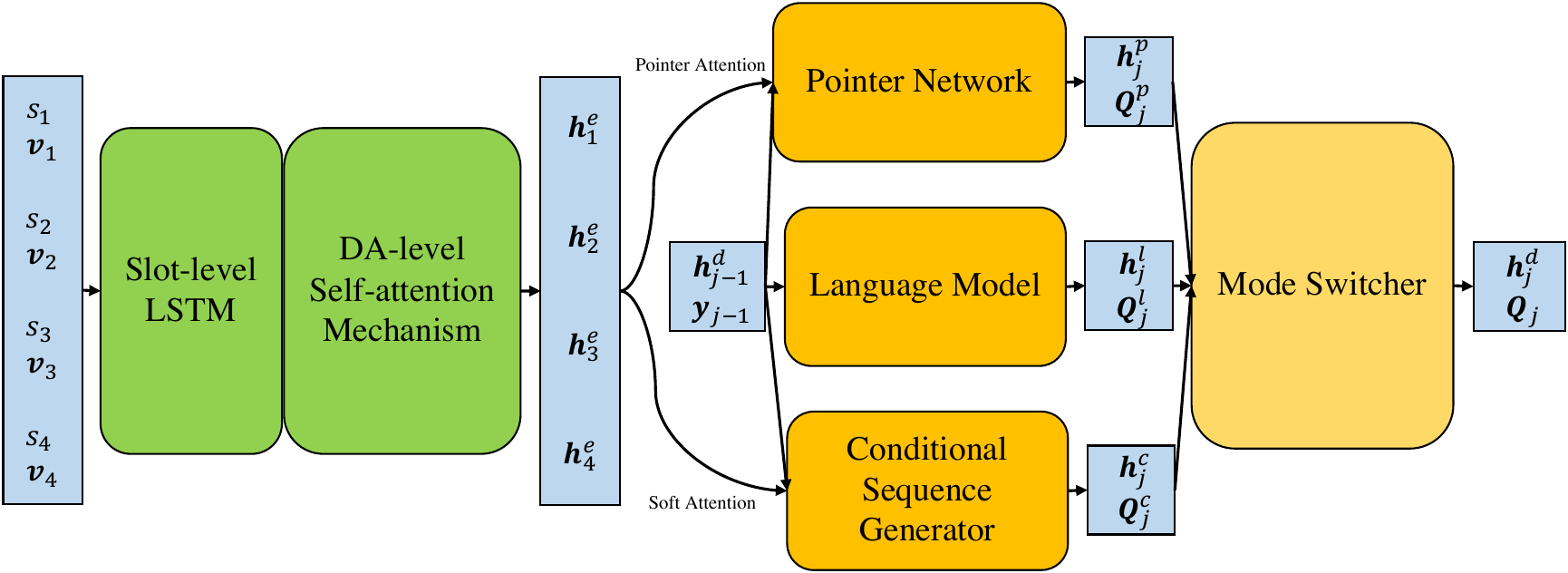}
		
		\caption{The architecture of the proposed model. Different modules are differentiated by colors.}
		\label{fig:Architecture}
	\end{figure*}
	
	In this work, we present a novel framework, heterogeneous rendering machines (HRM), that is both end-to-end and interpretable. Through using HRM as the decoder, we can interpret the mapping from input a DA to an utterance. HRM consists of a renderer set and a mode switcher. The renderer set is a collection of multiple decoders that vary in structure and functionality. In implementation, we set a conditional sequence generator~\citep{sutskever2014sequence,bahdanau2014neural,li2020rewriter} to render indicative slots and reworded slots, a pointer network~\citep{vinyals2015pointer} to explicitly copy the values of delexicalizable slots, and a language model~\citep{mikolov2010recurrent,Li_Yao_Qin_Peng_Liu_Li_2020} to produce context-agnostic words. The mode switcher is a discrete latent variable that selects a proper decoder from the renderer set for every generation step. We explore using two techniques, Gumbel-softmax~\citep{jang2016categorical} and VQ-VAE~\citep{van2017neural}, for the selection. Furthermore, a hierarchical encoder is adopted to represent DA as a collection of coarse-grained semantic units. We use a slot-level LSTM~\citep{hochreiter1997long} to encode slot values and a DA-level self-attention mechanism~\citep{vaswani2017attention} to capture the semantic correlations among them.
	
	Extensive experiments are conducted on multiple benchmark datasets. In terms of automatic metrics, our model is competitive with the existing state-of-the-art method. Qualitative analysis shows that our model is capable of interpreting the rendering process well. We also conduct human evaluation to confirm the effectiveness of our method.
	
\section{Architecture}
	
	Figure \ref{fig:Architecture} demonstrates our model. Firstly, a hierarchical encoder embeds input DA into a set of semantic units. Then, for each generation iteration, mode switcher aggregates the predictions made by different renderers.
	
	Formally, we represent an input DA as a $n$-sized set $\mathbf{x} = \{(s_1, \mathbf{v}_1), (s_2, \mathbf{v}_2), \cdots, (s_{n}, \mathbf{v}_n)\}$. Here $s_1$ is the act type and $v_1$ is a dummy symbol ``TYPE''. Other elements $(s_i, \mathbf{v}_i), i > 1$ are the slot value pairs. The ground truth utterance is a $m$-length list $\mathbf{y} = [y_1, y_2, \cdots, y_m]$. Every item in it is either a word or a phrase (i.e., slot value). For instance, the case in Figure \ref{fig:Data Case} is represented as
	\begin{equation} \nonumber
	\begin{aligned} 
	\mathbf{x} = \{ & (\mathrm{inform}, [\mathrm{TYPE}]), (\mathrm{name}, [\mathrm{Blue}, \mathrm{Spice}]), \\
	& (\mathrm{familyFriendly}, [\mathrm{yes}]), \\
	&  (\mathrm{priceRange}, [\mathrm{less}, \mathrm{than}, \mathrm{20}]) \} \\
	\mathbf{y} = [& ``\mathrm{The}", ``\mathrm{Blue~Spice}",`` \mathrm{is}", ``\mathrm{a}", ``\mathrm{low}", \\
	& ``\mathrm{cost}", ``\mathrm{venue}", ``\mathrm{.}", ``\mathrm{it's}",``\mathrm{a}", \\ 
	& ``\mathrm{family}", ``\mathrm{friendly}", ``\mathrm{location}", ``\mathrm{.}"]
	\end{aligned}.
	\end{equation}
	
	Originally, the output utterance is a sequence of words. The above format is obtained by merging successive words that match some input slot value (delexicalizable or reworded) verbatim into a phrase.
	
\subsection{Hierarchical Encoder}
	
	\begin{figure*}
		\centering
		\includegraphics[width=0.8\textwidth]{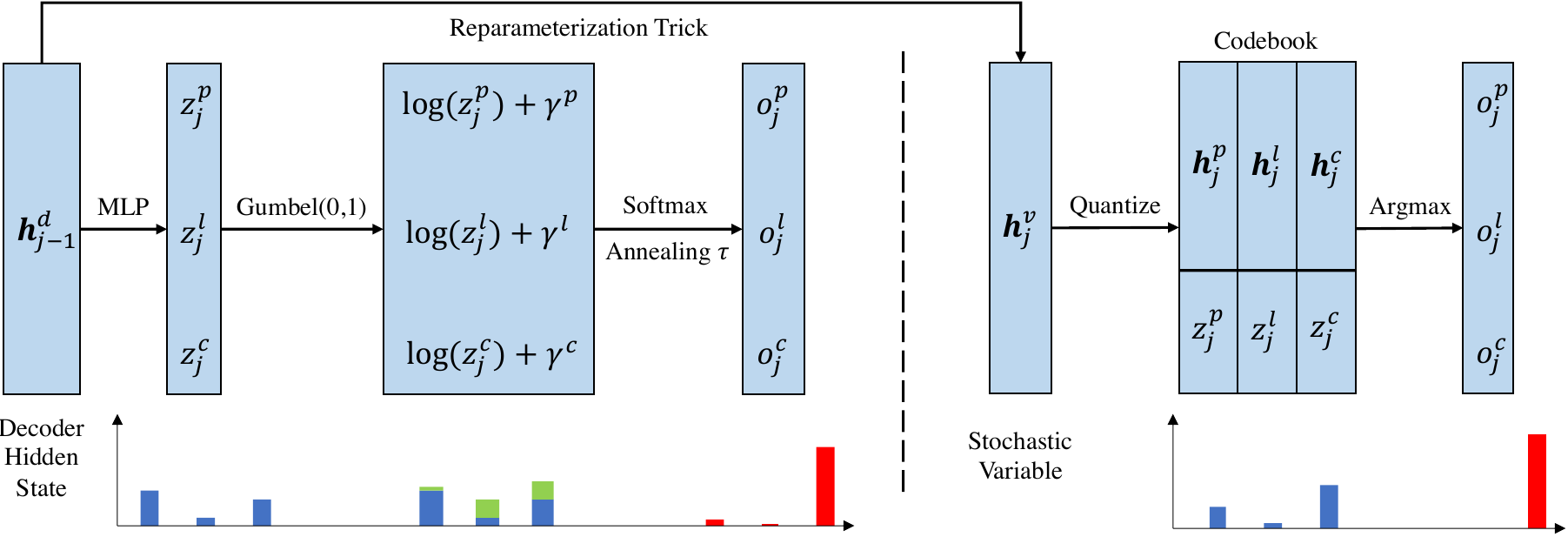}
		
		\caption{Two attempts to implement the mode switcher: Gumbel-softmax (left) and VQ-VAE (right).}
		\label{fig:Mode Switcher}
	\end{figure*}
	
	Hierarchical encoder converts input DA into a collection of coarse-grained semantic units.
	
	Firstly, a slot-level LSTM $f^{s}$ is used to encode each value $\mathbf{v}_i = [v_{i, 1}, v_{i, 2}, \cdots, v_{i, l_i}]$ as
	\begin{equation}
	\left\{\begin{aligned}
	\overrightarrow{\mathbf{h}}_{i,k}^v & = \overrightarrow{f}^s(\overrightarrow{\mathbf{h}}_{i,k-1}^v, \mathbf{E}(v_{i, k})) \\
	\overleftarrow{\mathbf{h}}_{i,k}^v & = \overleftarrow{f}^s(\overleftarrow{\mathbf{h}}_{i,k+1}^v, \mathbf{E}(v_{i, k})) \\ 
	\mathbf{h}_{i,k}^v & = \overrightarrow{\mathbf{h}}_{i,k}^v \oplus \overleftarrow{\mathbf{h}}_{i,k}^v
	\end{aligned}\right.,
	\end{equation}
	in which $\mathbf{E}$ is the word embedding. 
	
	Then, we embed every slots $s_i$ in DA $\mathbf{x}$:
	\begin{equation}
	\mathbf{h}^s_i = \mathbf{E}(s_i).
	\end{equation}
	
	We represent each slot value pair by column-wise vector concatenation as $\mathbf{h}^d_i = \mathbf{h}^s_i \oplus \mathbf{h}_{i,l_i}^v$.  In training, a word is randomly masked with ``UNK" with probability $1 / (1 + p)$ ($p$ is its frequency in the training set). This helps generalize the models to unseen values in the test set.
	
	Eventually, we leverage DA-level self-attention mechanism~\citep{vaswani2017attention} to capture the semantic correlations among slot values as 
	\begin{equation}
	\left\{\begin{aligned}
	\mathbf{H}^d & = [\mathbf{h}_1^d; \mathbf{h}_2^d; \cdots; \mathbf{h}_n^d] \\
	\mathbf{Q}^d, \mathbf{K}^d, \mathbf{V}^d & = \mathbf{W}_q\mathbf{H}^d, \mathbf{W}_k\mathbf{H}^d, \mathbf{W}_v\mathbf{H}^d \\
	\mathbf{H}^e & = \mathrm{Softmax}(\frac{\mathbf{Q}^d(\mathbf{K}^d)^T}{\sqrt{d}})\mathbf{V}^d \\
	\mathbf{H}^e & = [\mathbf{h}_1^e; \mathbf{h}_2^e; \cdots; \mathbf{h}_n^e]
	\end{aligned}\right.,
	\end{equation}
	where $[;]$ means row-wise vector concatenation and $d$ is the dimension of hidden layers.
	
	Prior methods mostly adopt LSTM or one-hot encoding as encoders. However, LSTM~\citep{hochreiter1997long} is order-sensitive and reads DA sequentially, ignoring the logic structure. One-hot encoding can't handle reworded slots. Our encoder circumvents all these issues.
	
\subsection{Heterogeneous Rendering Machines} 
\label{sec:Heterogeneous Rendering Machines}
	
	HRM is the decoder of our proposed model, which aims to interpret how neural generators render input DA into an utterance. It consists of two components, including a renderer set and a mode switcher.
	
\subsubsection{Renderer Set}
	
	Renderer set is a collection of decoders varying in both structure and functionality. In the proposed model, we set a pointer network, a conditional sequence generator, and a language model.
	
	In this section, we assume that it's the $j$-th generation step. The decoder hidden state and the generated item of prior step are denoted as $\mathbf{h}_{j-1}^d$ and $\mathbf{y}_{j-1}$, respectively. Note that $\mathbf{y}_{j-1}$ is embedded as $\mathbf{h}_i^s$ if it's a phrase directly copied from the value $\mathbf{v}_i$.

	\textit{Pointer Network} explicitly copies the slot value that exactly matches successive words in the utterance. Firstly, an LSTM $g^{p}$ is used to update the hidden state:
	\begin{equation}
	\mathbf{h}_j^p = g^p(\mathbf{h}_{j-1}^d, \mathbf{E}(y_{j-1})).
	\end{equation}
	Then, analogous to pointer network, we apply soft attention over input slot value pairs as
	\begin{equation}
	\left\{\begin{aligned}
	\beta_{j,i} & = \mathbf{v}^T_{\beta} \tanh(\mathbf{W}_{\beta} \mathbf{h}^p_{j} + \mathbf{U}_{\beta}\mathbf{h}^e_i) \\
	\mathbf{Q}^p_j & = \mathrm{Softmax}([\beta_{j, 1}, \beta_{j, 2}, \cdots, \beta_{j, n}])
	\end{aligned}\right.,
	\end{equation}
	where $\mathbf{v}_{\beta}$, $\mathbf{W}_{\beta}$, and $\mathbf{U}_{\beta}$ are learnable.  
	
	\textit{Conditional Sequence Generator} renders some slot value into various forms in the sentence. Firstly, we apply attention mechanism over the encoded DA representations:
	\begin{equation}
	\left\{\begin{aligned}
	\alpha_{j,i} & = \mathbf{v}^T_{\alpha} \tanh(\mathbf{W}_{\alpha} \mathbf{h}^d_{j - 1} + \mathbf{U}_{\alpha}\mathbf{h}^e_i) \\
	\alpha'_{j,i} & = \frac{\exp(\alpha_{j,i})}{\sum_{1 \le k \le n} \exp(\alpha_{j,k})} \\
	\mathbf{h}^a_j & = \sum_{1 \le i \le n} \alpha'_{j,i} \mathbf{h}_i^e
	\end{aligned}\right..
	\end{equation} 	
	Then, we employ a LSTM $g^c$ to read prior item and update the decoder hidden state as
	\begin{equation}
	\mathbf{h}_j^c = g^c(\mathbf{h}_{j-1}^d, \mathbf{E}(y_{j-1}) \oplus \mathbf{h}^a_j).
	\end{equation}  
	Finally, it predicts the next word distribution as
	\begin{equation}
	\mathbf{Q}_j^c = \mathbf{W}_{c} (\mathbf{h}^c_j \oplus \mathbf{h}^a_j),
	\end{equation}
	where $\mathbf{W}_{c}$ is also a learnable parameter.
	
	\textit{Unconditional Language Modeling} produces context-agnostic words (e.g., ``is" and ``a") in Figure \ref{fig:Data Case}. Firstly, we apply an LSTM $g^l$ to read the previous item as 
	\begin{equation}
	\mathbf{h}_j^l = g^l(\mathbf{h}_{j-1}^d, \mathbf{E}(y_{j-1})).
	\end{equation}
	Then, we predict the next word distribution:
	\begin{equation}
	\mathbf{Q}^l_j = \mathrm{Softmax}(\mathbf{W}_l \mathbf{h}_j^l).
	\end{equation}
	
	$\mathbf{Q}^c_j$ and $\mathbf{Q}^l_j$ are the distributions over predefined word vocabulary, while $\mathbf{Q}^p_j$ is the one over input slot values.
	
\subsubsection{Mode Switcher} 

	Mode switcher is a discrete latent variable. At each generation step, its aim is to choose an appropriate decoder from the renderer set. To implement it, we explore using Gumbel-softmax trick and further introduce a variant of VQ-VAE.
	
	The output $\mathbf{o}_j = [o_j^p, o_j^c, o_j^l]$ of it is an one-hot vector. We use it to aggregate the updated hidden states and the predicted next word distributions of different renderers:
	\begin{equation}
	\label{equ:Equation 11}
	\left\{\begin{aligned}
	\mathbf{h}^d_j & = o_j^p\mathbf{h}_j^p + o_j^c\mathbf{h}_j^c + o_j^l\mathbf{h}_j^l \\
	\mathbf{Q}_j & = o_j^p\mathbf{Q}_j^p + o_j^c\mathbf{Q}_j^c + o_j^l\mathbf{Q}_j^l
	\end{aligned}\right..
	\end{equation}
	
	At test time, we make the next word prediction as $\hat{y}_j = \mathop{\arg\max} \mathbf{Q}_j$.  In the rest of this section, we describe two approaches to implement it.
	
	\textit{Gumbel-softmax Trick}~\citep{jang2016categorical} solve the nondifferentiability problem of sampling from a categorical distribution. The main idea is using the differentiable sample from a gumbel-softmax distribution which can be smoothly annealed into given categorical distribution.
	
	The procedure is demonstrated in the left part of Figure \ref{fig:Mode Switcher}. Firstly, we predict a categorical distribution $\mathbf{z}_j$ over the renderer set as
	\begin{equation}
	\label{equ: Equation 12} 
	\left\{\begin{aligned}
	\mathrm{[} \theta_j^p, \theta_j^c, \theta_j^l ] & = \mathbf{W}_\theta\tanh(\mathbf{U}_\theta\mathbf{h}^d_{j-1}) \\
	[ z_j^p, z_j^c, z_j^l ] & = \mathrm{Softmax}([ \theta_j^p, \theta_j^c, \theta_j^l ])
	\end{aligned}\right..
	\end{equation} 
	
	Then, the following operation is used as a differentiable approximation to sampling: 
	\begin{equation}
	o_j^r = \frac{\exp((\log(z_j^r) + \gamma^r) / \tau)}{\sum_{r' \in \{p, c, l\}} \exp((\log(z_j^{r'}) + \gamma^{r'}) / \tau)},
	\end{equation}
	where subscript $r$ iterates over $\{p, c, l\}$. $\gamma^p$, $\gamma^c$, and $\gamma^l$ are i.i.d  samples drawn from $\mathrm{Gumbel}(0, 1)$~\citep{gumbel1948statistical}.
	
	\textit{VQ-VAE}~\citep{van2017neural} use vector-quantized variational autoencoder (VQ-VAE) to learn discrete representations. The recognition network outputs discrete codes and its parameters are optimized by straight-through estimator~\citep{bengio2013estimating}. In implementations, we have made two changes: 1) reparameterization trick~\citep{kingma2013auto} is used to add randomness; 2) the codebook (i.e., embedding space in the original paper) is dynamic rather than static. 
	
	As depicted in the right part of Figure \ref{fig:Mode Switcher}, firstly, we apply reparameterization trick on hidden state $\mathbf{h}^d_{j-1}$ to get a stochastic variable $\mathbf{h}^v_j$ as
	\begin{equation}
	\left\{\begin{aligned}			 
	\mu_j & = \mathbf{W}_{\mu}\tanh(\mathbf{U}_{\mu}\mathbf{h}^d_{j-1}) \\
	\sigma_j & = \mathbf{W}_{\sigma}\tanh(\mathbf{U}_{\sigma}\mathbf{h}^d_{j-1}) \\
	\mathbf{h}^v_j & = \mu_j + \epsilon \odot \exp(\sigma_j)
	\end{aligned}\right.,
	\end{equation}
	where $\epsilon$ is sampled from standard Gaussian distribution and $\odot$ is element-wise product.
	
	Then, we quantize the variable $\mathbf{h}^v_j$ in terms of the dynamic codebook that contains different decoder hidden states:
	\begin{equation} 
	\left\{\begin{aligned}	
	z_j^r & = \| \mathbf{h}^v_j  - \mathbf{h}^r_j  \|_2, r \in \{p, c, l\} \\
	r' & = \mathop{\arg\min}_{r \in \{p, c, l\}} z_j^r \\
	[ o_j^p, o_j^c, o_j^l ] & = \mathrm{OneHot}(r')
	\end{aligned}\right.,
	\end{equation} 
	where $\|\|_2$ is Euclidian distance. Function $\mathrm{OneHot}$ constructs a $3$-length one-hot vector, where the $r'$-th value is $1$. Following VQ-VAE, we use straight-through estimator to approximate the gradient of $\arg\min$.
	
	\subsection{Discussion}
	\label{sec:Discussion}
	
	In this section, we show the rationality of HRM and how it interprets neural generators.
	
	\paragraph{Rationality.} We utilize probability theory to analyze the essence of HRM. Generally, our decoder decomposes the joint probability of producing a natural language sentence $\mathbf{y}$ into the ordered conditionals:
	\begin{equation}
	P(\mathbf{y} | \mathbf{x}) = \prod_{1 \le i \le m} P(y_i | y_{<i}, \mathbf{x}).
	\end{equation}  
	
	Since an item is produced by either copying, conditional generation, or context-agnostic generation, its probability can be decomposed as
	\begin{equation} 
	\label{equ:Equation 17}
	\begin{aligned}
	& P(y_i | y_{<i}, \mathbf{x})  = \\ 
	\sum_{r \in \{p, c, l\}} & P(y_i | y_{<i}, \mathbf{x}, r) P(r | y_{<i}, \mathbf{x})
	\end{aligned},
	\end{equation} 
	which corresponds to Equation \ref{equ:Equation 11}. $P(r | y_{<i}, \mathbf{x}), r \in \{p, c, l\}$ is the output of mode switcher. $P(y_i | y_{<i}, \mathbf{x}, r)$ is the next item distribution predicted by corresponding renderer. For example, $l$ represents language model.
	
	\begin{table*}
		\centering
		
		\setlength{\tabcolsep}{1.8mm}{}
		\begin{tabular}{c|c|cc|cc|cc|cc|c}
			\hline
			\multicolumn{2}{c|}{\multirow{2}{*}{Model}} & \multicolumn{2}{c|}{Restaurant}  &  \multicolumn{2}{c|}{Hotel} & \multicolumn{2}{c|}{Laptop} & \multicolumn{2}{c|}{Television} &  \multicolumn{1}{c}{E2E-NLG}  \\
			
			\multicolumn{2}{c|}{} & BLEU & ERR & BLEU & ERR & BLEU & ERR & BLEU & ERR & BLEU \\
			
			\hline
			\multicolumn{2}{c|}{HLSTM~\citep{wen-etal-2015-stochastic}} & $0.747$ & $0.74$ & $0.850$ & $2.67$ & $0.513$ & $1.10$ & $0.525$ & $2.50$ & - \\
			
			\multicolumn{2}{c|}{SCLSTM~\citep{wen-etal-2015-semantically}} & $0.753$ & $0.38$ & $0.848$ & $3.07$ & $0.512$ & $0.79$ & $0.527$ & $2.31$ & - \\
			
			\multicolumn{2}{c|}{TGen~\citep{dusek-jurcicek-2016-sequence}} & - & - & - & - & - & - & -  & - &  $0.659$ \\
			
			\multicolumn{2}{c|}{RALSTM~\citep{tran-nguyen-2017-natural}}  & $0.779$ & $\mathbf{0.16}$ & $0.898$ & $\mathbf{0.43}$ & $0.525$ & $0.42$ & $0.541$ & $0.63$ & - \\
			
			\cdashline{1-11}
			\multicolumn{2}{c|}{NLG-LM~\citep{zhu2019multi}}  & $0.795$ & - & $\mathbf{0.939}$ & - & $0.586$ & - & $\mathbf{0.617}$ & - & $0.684$ \\
			
			\hline
			
			\multirow{3}{*}{Our Model} & Gumbel-softmax & $0.776$ & $0.21$ & $0.903$ & $0.77$ & $0.561$ & $0.63$ & $0.581$ & $0.79$ & $0.667$ \\
			
			& VQ-VAE & $0.789$ & $0.18$ & $0.921$ & $0.55$ & $0.554$ & $0.65$ & $0.598$ & $0.70$ & $0.681$ \\
			
			\cdashline{2-11}
			& Softmax & $\mathbf{0.812}$ & $0.20$ & $0.923$ & $0.46$ & $\mathbf{0.591}$ & $\mathbf{0.31}$ & $0.610$ & $\mathbf{0.51}$ & $\mathbf{0.697}$ \\
			
			\hline
		\end{tabular}
		\caption{Experiment results on five datasets for all baselines and our models.} 
		
		\label{tab:Main Experiment}
	\end{table*}
	
	To permit interpretability, mode switcher is modeled as a discrete latent variable. In Equation \ref{equ:Equation 17}, it means that distribution $P(r | y_{<i}, \mathbf{x})$ is ``one hot". However, experiments show that this degrades the performances.
	
	\paragraph{Interpretability.} We can understand how neural generators render input DA into an utterance using the following procedure. For each item $y_j$ in the generated sentence, the output $\mathbf{o}_j$ of mode switcher indicates which renderer produces it. If the renderer is pointer network, pointer attention $\mathbf{Q}_j^p$ shows where the item is copied from. If it's conditional sequence generator, soft attention $[\alpha'_{j,1}, \alpha'_{j,1}, \cdots, \alpha'_{j,n}]$ indicates which slot value the item is reworded from. Otherwise, the item is a context-agnostic word generated by the language model.
	
\section{Training Criterions}
	
	The following cross entropy loss is incurred for training:
	\begin{equation} 
	L^c = \sum_{1 \le j \le m} -\log \mathbf{Q}_j[y_j].
	\end{equation}
	
	\begin{table}[t]
		\centering
		
		\setlength{\tabcolsep}{1.4mm}{}
		\begin{tabular}{c|ccccc}
			\hline
			
			& R & H & L & T & E \\
			
			\hline
			$|$training set$|$  & $3114$ & $3223$ & $7944$ & $4221$ & $42061$ \\
			
			$|$validation set$|$ & $1039$ & $1075$ & $2649$ & $1407$ & $4672$ \\
			
			$|$test set$|$ & $1039$ & $1075$ & $2649$ & $1407$ & $4693$ \\
			
			total & $5192$ & $5373$ & $13242$ & $7035$ & $47366$ \\
			
			\hline
			
			DA types & $8$ & $8$ & $14$ & $14$ & $1$ \\
			slot types & $12$ & $12$ & $20$ & $16$ & $8$ \\	
			
			\hline
		\end{tabular}
		\caption{The details of different datasets.}
		
		\label{tab:Dataset Detail}
	\end{table}
	
	If mode switcher is implemented as VQ-VAE, we add two extra criterions. Firstly, following \citet{van2017neural}, we have:
	\begin{equation}
	L^d =  \sum_{1 \le j \le m}  \|\mathrm{sg}[\mathbf{h}^v_j] - \mathbf{h}_j^{r'}\|_2^2 +  
	\rho  \|\mathbf{h}^v_j - \mathrm{sg}[\mathbf{h}_j^{r'}]\|_2^2,
	\end{equation} 
	where $\mathrm{sg}$ stands for the stopgradient operator that is defined as identity at forward computation time
	and has zero partial derivatives. In all experiments, we set $\rho$ as $0.25$. Secondly, following \citet{kingma2013auto}, we adopt KL divergence to avoid posterior collapse:
	\begin{equation}
	L^v = \sum_{1 \le j \le m} \left( \mu^2_j + \exp(\sigma_j) - (1 + \sigma_j)\right).
	\end{equation} 
	If using Gumbel-softmax, the annealing schedule for $\tau$ is the same as that in~\citet{jang2016categorical}.

\section{Experiments}
	
	To verify the effectiveness of our model, we have conducted comprehensive studies on five datasets. The main experiments show that our model is competitive with the current state-of-the-art approach. Case studies demonstrate that the proposed model can interpret the rendering process well. we also investigate ablation experiments to explore the impacts of some components.
	
\subsection{Settings} 
	
	We evaluate the models on five benchmark datasets. The Hotel dataset and the Restaurant dataset are collected in~\citep{wen-etal-2015-stochastic}. The Laptop dataset and the TV dataset are from~\citep{wen-etal-2015-semantically}. The E2E-NLG dataset is released by a shared challenge~\citep{novikova-etal-2017-e2e}\footnote{http://www.macs.hw.ac.uk/InteractionLab/E2E/.}. All the datasets used in our paper follow the same format, pretreatment, and partition as in~\citep{wen-etal-2015-stochastic,wen-etal-2015-semantically,novikova-etal-2017-e2e}. Other details of the datasets are demonstrated in Table \ref{tab:Dataset Detail}. 
	
	For fairness, we use the official evaluation scripts from repositorys:  E2E-NLG\footnote{https://github.com/tuetschek/e2e-metrics.} and  RNN-LG\footnote{https://github.com/shawnwun/RNNLG.}. The automatic metrics include BLEU and slot error rate (ERR). ERR is computed as $\mathrm{ERR} = \frac{p + q}{N}$,
	where $N$ is the total number of slots in the DA, and $p$, $q$ is the amount of missing and redundant slots in the generated utterance, respectively.
	
	\begin{figure*}
		\centering
		
		\includegraphics[width=0.90\textwidth]{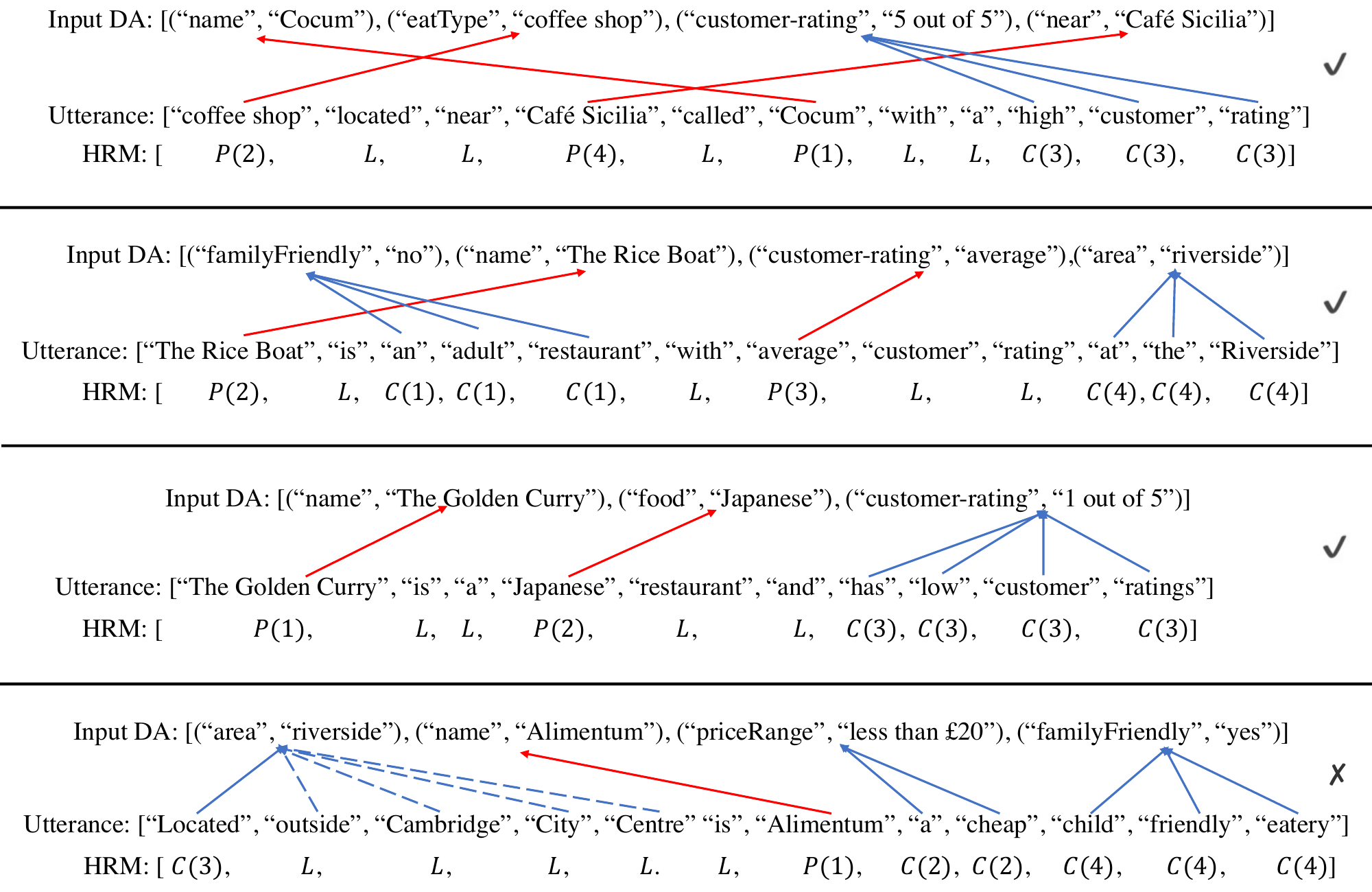}
		
		\caption{The cases are extracted from E2E-NLG dataset to show the interpretability of HRM.}
		\label{fig:Cases about Interpretability}
	\end{figure*}
	
	We adopt the same configurations for all five datasets. The dimensionalities for all embeddings are $256$. The hidden units of all layers are set as $512$. We adopt $3$ layers of self-attention and each of them has $4$ heads. L2 regularization is set as $1 \times 10^{-6}$ and the dropout ratio is assigned $0.4$ for reducing overfit. Above setting is obtained by using grid search. We use Adam~\citep{kingma2014adam} to optimize model parameters. All the studies are conducted at GeForce RTX 2080T. For each DA, we over-generate $10$ utterances through beam search and select the top $5$ candidates. In experiments, we select the model that works the best on the dev set, and then evaluate it on the test set. The improvements of our models over the baselines are statistically significant with $p < 0.05$ under t-test.
	
\subsection{Main Results}
	
	The baselines for comparison are as follows:
	
	\begin{itemize}
		
		\item HLSTM~\citep{wen-etal-2015-stochastic} designs a heuristic gate to control the states of slot values, guaranteeing that all of them are accurately captured;
		
		\item SCLSTM~\citep{wen-etal-2015-semantically} introduces an extra ``reading gate” to improve standard LSTM;
		
		\item TGen~\citep{dusek-jurcicek-2016-sequence} uses encoder-decoder architecture augmented with attention mechanism to generate utterances from input DA;
		
		\item RALSTM~\citep{tran-nguyen-2017-natural} presents a RNN-based decoder to select and aggregate the semantic elements produced by attention mechanism;
		
		\item NLG-LM~\citet{zhu2019multi} incorporates language modeling into response generation. In particular, it provides current state-of-the-art performances on BLEU score for all the datasets.
		
	\end{itemize}
	
	We follow the baseline results as reported in \citet{tran-nguyen-2017-natural,zhu2019multi}.
	
	Table \ref{tab:Main Experiment} presents main results. First of all, VQ-VAE is more effective than gumbel-softmax for modeling mode switcher. Except for Laptop dataset, using VQ-VAE obtains better performances than using gumbel-softmax. In terms of BLEU scores, it improves the performances by $1.68\%$ on Restaurant dataset, $1.99\%$ on Hotel dataset, $2.84\%$ on Television dataset, and $2.06\%$ on E2E-NLG dataset. For correctness, it also reduces the ERR by $14.29\%$ on Restaurant dataset, $28.57\%$ on Hotel dataset, and $11.39\%$ on Television dataset. Secondly, our model is not only interpretable but also competitive with current state-of-the-art method. For instance, our BLEU scores are lower than those of NLG-LM by only $0.75\%$ on Restaurant dataset and $0.44\%$ on E2E-NLG dataset.
	
	Lastly, we discover that there is a trade-off between Interpretability and effectiveness. The results shown in the last row are obtained by directly using the categorical distribution $[ z_j^p, z_j^c, z_j^l ]$ predicted in Equation \ref{equ: Equation 12} as the output $\mathbf{o}_j$ of mode switcher. Although it degrades the interpretability, the performances are notably improved. Compared with NLG-LM, it achieves the increasements on BLEU scores of $2.14\%$ on Restaurant dataset, $0.85\%$ on Laptop dataset, and  $1.90\%$ on E2E-NLG dataset. For ERR, the scores are also reduced by $26.19\%$ on Laptop dataset and $19.05\%$ on Television dataset. We guess that using discrete representations limits the flexibility of neural networks.
	
\subsection{Case Study}

	In this section, we show that HRM is capable of interpreting the rendering process well.
	
	Using the procedure described in Section \ref{sec:Discussion}, we label a generated item as $P(i)$ if it's copied from the $i$-th slot value, $C(i)$ if it's reworded from the $i$-th slot value, or $L$ otherwise. Several examples are depicted in Figure \ref{fig:Cases about Interpretability}, where blue lines and red lines denote paraphrasing and copying, respectively. In most of the cases, HRM is of great interpretability. For example, in the third case, phrase ``has low customer ratings" is aligned with value ``1 out of 5". In the second case, phrase ``an adult restaurant" is aligned with slot value (``familyFriendly", ``no"). There also exists a few mistakes. For instance, in the fourth case, phrase ``outside Cambridge City Centre" is not tied with value ``riverside". We attribute this to the high fault tolerance of end-to-end neural networks. In addition, here we omit act type for brevity.
	
\subsection{Ablation Study}
	
	\begin{table}[t]
		\centering

		\begin{tabular}{c|cc}
			\hline
			
			Method & Hotel & E2E-NLG  \\
			
			\hline
			Our Model & $\mathbf{0.921}$ & $\mathbf{0.681}$ \\ 
			
			\hline
			w/o Self-attention & $0.908$ & $0.651$  \\
			
			w/o Self-attention, w/ LSTM & $0.914$ & $0.672$ \\
			
			\cdashline{1-3}
			w/o Pointer Net & $0.862$ & $0.665$ \\
			w/o Pointer Net, w/ Delex. & $0.910$ & $0.669$ \\
			
			\hline
		\end{tabular}
		\caption{Ablation experiments on two datasets.}
		
		\label{tab:Ablation Study}
	\end{table} 
	
	As shown in Table \ref{tab:Ablation Study}, we conduct ablation study to explore the impacts of components.
	
	\paragraph{Effect of DA-level Encoder.} By removing self-attention, the BLEU scores sharply drop by $1.41\%$ and $4.41\%$. Hence it's useful to capture the semantic correlations among slot values. Additionaly, replacing self-attention with LSTM decreases the BLEU scores by $0.76\%$ and $1.32\%$. We credit this to the order-insensitivity of self-attention.
	
	\paragraph{Effect of Copying Mechanism.} By removing pointer network, the BLEU scores decrease by $6.41\%$ and $2.35\%$. Hence directly copying words from input DA is useful. Replacing pointer network with delexicalization reduces the BLEU scores by $1.19\%$ and $1.76\%$. Besides, delexicalization can only apply to delexicalizable slots. These indicate that pointer network is more effective.
	
\subsection{Human Evaluation}
	
	\begin{table}[t]
		\centering
		\begin{tabular}{c|c}
			\hline
			
			Method & E2E-NLG  \\
			
			\hline
			TGen & $0.562$ \\
			
			NLG-LM & $0.495$ \\
			
			\hline
			Our Model w/ Gumbel-softmax & $0.815$ \\
			
			Our Model w/ VQ-VAE & $0.872$ \\
			
			Our Model w/ Softmax & $0.621$  \\
			
			\hline
		\end{tabular}
		\caption{Human evaluation on E2E-NLG dataset.}
		
		\label{tab:Human Evaluation}
	\end{table} 
	
	We conduct human evaluation to quantitively make comparisons on interpretability among different models. The task consists of three stages. Firstly, we introduce an alignment score to assess the interpretability of a model. It's computed as $p / N$. $p$ is the number of slot values that are correctly aligned with the utterance by a model (see Figure \ref{fig:Cases about Interpretability}). $N$ is the total amount of slot values. Then, we define how various generators interpret their rendering processes. For our models, we do interpretation as described in Section \ref{sec:Discussion}. For the baselines (e.g., TGen), we use the computed attention weights to align the input DA with the generated sentence. Eventually, we randomly sample $200$ cases from the test set as the benchmark set and recruit 15 people to judge whether a slot value is aligned well with the generated utterance (i.e., getting $p$). The designed task is very simple and highly consistent among workers.
	
	Table \ref{tab:Human Evaluation} demonstrates the experiment results. From it, we can draw the following conclusions. Firstly, better performances may lead to worse interpretability. For example, the state-of-the-art model, NLG-LM, underperforms a simple baseline, TGen, by $13.5\%$. From Table \ref{tab:Main Experiment}, we can see that, for our model, using softmax generally obtains higher F1 scores than adopting discrete latent variable models. However, here we find the interpretability score of using softmax is lower than using VQ-VAE or Gumbel-softmax. Secondly, our models consistently and significantly outperform prior baselines. For example, the F1 score of Our Mode w/ VQ-VAE outnumbers TGen by $35.6\%$ and NLG-LM by $43.2\%$. Thirdly, VQ-VAE is better than Gumbel-softmax in terms of both BLEU score and interpretability score. For example, in Table \ref{tab:Main Experiment}, using VQ-VAE outperforms using Gumbel-softmax by $1.65\%$ on Restaurant, $2.84\%$ on Television, and $2.06\%$ on E2E-NLG. In Table \ref{tab:Human Evaluation}, the increase percentage of interpretability score is $6.54\%$.

\section{Related Work}
	
	In task-oriented dialogue systems, NLG is the final module to produce user-facing system utterances, which is directly related to the perceived quality and usability. Traditional approaches generally divide the NLG task into a pipeline of sentence planning and surface realization~\citep{oh2000stochastic,ratnaparkhi2000trainable,mirkovic2011dialogue,cheyer2014method}. Sentence planning first converts an input DA into a tree-like structure, and then surface realization maps the intermediate structure into the final surface form. For example, \citet{oh2000stochastic} use a class-based n-gram language model and a template-based reranker. \citet{ratnaparkhi2000trainable} address the limitations of n-gram language models by using more complex syntactic trees. \citet{mairesse2014stochastic} employ a phrase-based generator that learns from a semantically aligned corpus. Although these methods are adequate and of great interpretability, they are heavily dependent on handcraft rules and expert knowledge. Moreover, the sentences generated from rule-based systems are often rigid, without the diversity and naturalness of human language.
	
	Lately, there is a surge of interest in utilizing neural networks to build corpus-based NLG models~\citep{wen-etal-2015-stochastic,dusek-jurcicek-2016-sequence,tran-nguyen-2017-natural,li-etal-2020-slot}. The main superiority is facilitating end-to-end training on the unaligned corpus. For example, \citet{wen-etal-2015-stochastic} present a heuristic gate to guarantee that all slot value pairs are accurately captured during generation. \citet{wen-etal-2015-semantically} introduce a novel SC-LSTM with an additional reading cell to learn gated mechanism and language model jointly.  \citet{dusek-jurcicek-2016-sequence} use encoder-decoder architecture augmented with attention mechanism to generate utterances from input DA. \citet{tran-nguyen-2017-natural} use a RNN-based decoder to select and aggregate the semantic elements produced by attention mechanism. Most recently, \citet{zhu2019multi}  incorporate a language model task into the response generation process to boost the naturalness of generated utterances. \citet{li-etal-2020-slot} study the slot consistency issue and propose a novel iterative rectification network to address it. While plenty of state-of-the-art performances have been obtained, they are all treated as black boxes, and thus lack interpretability. Delexicalization~\citep{wen-etal-2015-semantically,tran-nguyen-2017-natural,li-etal-2020-handling} to some extent raises the interpretability as it directly locates the position of slot values in the utterance. Nevertheless, it is applicable for delexicalizable slots only. In E2E-NLG dataset, most of the slots are reworded or indicative. \citet{nayak2017plan} also observe that using delexicalization results in mistakes. 
	
\section{Conclusion}
	
	In this paper, we present heterogeneous rendering machines (HRM) to improve the interpretability of NLG models. It consists of a renderer set and a mode switcher. The renderer set contains multiple decoders that vary in structure and functionality. The mode switcher is a discrete latent variable that chooses an appropriate decoder from the renderer set in every generation step. Extensive experiments have been conducted on five datasets, demonstrating that our model is competitive with the current state-of-the-art method. Qualitative studies show that our model can interpret the rendering process well. Human evaluation further confirms its effectiveness in interpretability.
	
	Currently, a severe problem in interpretable NLG is lacking a proper evaluation metric. Mainstream metrics such as BLEU are not applicable. Using our alignment score demands massive annotation efforts. We will work hard on this issue in future research.
	
\section*{Acknowledgments}
	
	This work was supported by Ant Group through Ant Research Program. We thank anonymous reviewers for their valuable and constructive comments.
	
	\bibliography{aaai21}

\begin{thebibliography}{30}
\providecommand{\natexlab}[1]{#1}
\providecommand{\url}[1]{\texttt{#1}}
\providecommand{\urlprefix}{URL }
\expandafter\ifx\csname urlstyle\endcsname\relax
  \providecommand{\doi}[1]{doi:\discretionary{}{}{}#1}\else
  \providecommand{\doi}{doi:\discretionary{}{}{}\begingroup
  \urlstyle{rm}\Url}\fi

\bibitem[{Bahdanau, Cho, and Bengio(2014)}]{bahdanau2014neural}
Bahdanau, D.; Cho, K.; and Bengio, Y. 2014.
\newblock Neural machine translation by jointly learning to align and
  translate.
\newblock \emph{arXiv preprint arXiv:1409.0473} .

\bibitem[{Bengio, L{\'e}onard, and Courville(2013)}]{bengio2013estimating}
Bengio, Y.; L{\'e}onard, N.; and Courville, A. 2013.
\newblock Estimating or propagating gradients through stochastic neurons for
  conditional computation.
\newblock \emph{arXiv preprint arXiv:1308.3432} .

\bibitem[{Cheyer and Guzzoni(2014)}]{cheyer2014method}
Cheyer, A.; and Guzzoni, D. 2014.
\newblock Method and apparatus for building an intelligent automated assistant.
\newblock US Patent 8,677,377.

\bibitem[{Du{\v{s}}ek and
  Jur{\v{c}}{\'\i}{\v{c}}ek(2016)}]{dusek-jurcicek-2016-sequence}
Du{\v{s}}ek, O.; and Jur{\v{c}}{\'\i}{\v{c}}ek, F. 2016.
\newblock Sequence-to-Sequence Generation for Spoken Dialogue via Deep Syntax
  Trees and Strings.
\newblock In \emph{Proceedings of the 54th Annual Meeting of the Association
  for Computational Linguistics (Volume 2: Short Papers)}, 45--51. Berlin,
  Germany: Association for Computational Linguistics.
\newblock \doi{10.18653/v1/P16-2008}.
\newblock \urlprefix\url{https://www.aclweb.org/anthology/P16-2008}.

\bibitem[{Gumbel(1948)}]{gumbel1948statistical}
Gumbel, E.~J. 1948.
\newblock \emph{Statistical theory of extreme values and some practical
  applications: a series of lectures}, volume~33.
\newblock US Government Printing Office.

\bibitem[{Hochreiter and Schmidhuber(1997)}]{hochreiter1997long}
Hochreiter, S.; and Schmidhuber, J. 1997.
\newblock Long short-term memory.
\newblock \emph{Neural computation} 9(8): 1735--1780.

\bibitem[{Jain and Wallace(2019)}]{jain-wallace-2019-attention}
Jain, S.; and Wallace, B.~C. 2019.
\newblock {A}ttention is not {E}xplanation.
\newblock In \emph{Proceedings of the 2019 Conference of the North {A}merican
  Chapter of the Association for Computational Linguistics: Human Language
  Technologies, Volume 1 (Long and Short Papers)}, 3543--3556. Minneapolis,
  Minnesota: Association for Computational Linguistics.
\newblock \doi{10.18653/v1/N19-1357}.
\newblock \urlprefix\url{https://www.aclweb.org/anthology/N19-1357}.

\bibitem[{Jang, Gu, and Poole(2016)}]{jang2016categorical}
Jang, E.; Gu, S.; and Poole, B. 2016.
\newblock Categorical reparameterization with gumbel-softmax.
\newblock \emph{arXiv preprint arXiv:1611.01144} .

\bibitem[{Juraska et~al.(2018)Juraska, Karagiannis, Bowden, and
  Walker}]{juraska-etal-2018-deep}
Juraska, J.; Karagiannis, P.; Bowden, K.; and Walker, M. 2018.
\newblock A Deep Ensemble Model with Slot Alignment for Sequence-to-Sequence
  Natural Language Generation.
\newblock In \emph{Proceedings of the 2018 Conference of the North {A}merican
  Chapter of the Association for Computational Linguistics: Human Language
  Technologies, Volume 1 (Long Papers)}, 152--162. New Orleans, Louisiana:
  Association for Computational Linguistics.
\newblock \doi{10.18653/v1/N18-1014}.
\newblock \urlprefix\url{https://www.aclweb.org/anthology/N18-1014}.

\bibitem[{Kingma and Ba(2014)}]{kingma2014adam}
Kingma, D.~P.; and Ba, J. 2014.
\newblock Adam: A method for stochastic optimization.
\newblock \emph{arXiv preprint arXiv:1412.6980} .

\bibitem[{Kingma and Welling(2013)}]{kingma2013auto}
Kingma, D.~P.; and Welling, M. 2013.
\newblock Auto-encoding variational bayes.
\newblock \emph{arXiv preprint arXiv:1312.6114} .

\bibitem[{Li et~al.(2020{\natexlab{a}})Li, Li, Yao, and
  Li}]{li-etal-2020-handling}
Li, Y.; Li, H.; Yao, K.; and Li, X. 2020{\natexlab{a}}.
\newblock Handling Rare Entities for Neural Sequence Labeling.
\newblock In \emph{Proceedings of the 58th Annual Meeting of the Association
  for Computational Linguistics}, 6441--6451. Online: Association for
  Computational Linguistics.
\newblock \doi{10.18653/v1/2020.acl-main.574}.
\newblock \urlprefix\url{https://www.aclweb.org/anthology/2020.acl-main.574}.

\bibitem[{Li and Yao(2020)}]{li2020rewriter}
Li, Y.; and Yao, K. 2020.
\newblock Rewriter-Evaluator Framework for Neural Machine Translation.
\newblock \emph{arXiv preprint arXiv:2012.05414} .

\bibitem[{Li et~al.(2020{\natexlab{b}})Li, Yao, Qin, Che, Li, and
  Liu}]{li-etal-2020-slot}
Li, Y.; Yao, K.; Qin, L.; Che, W.; Li, X.; and Liu, T. 2020{\natexlab{b}}.
\newblock Slot-consistent {NLG} for Task-oriented Dialogue Systems with
  Iterative Rectification Network.
\newblock In \emph{Proceedings of the 58th Annual Meeting of the Association
  for Computational Linguistics}, 97--106. Online: Association for
  Computational Linguistics.
\newblock \doi{10.18653/v1/2020.acl-main.10}.
\newblock \urlprefix\url{https://www.aclweb.org/anthology/2020.acl-main.10}.

\bibitem[{Li et~al.(2020{\natexlab{c}})Li, Yao, Qin, Peng, Liu, and
  Li}]{Li_Yao_Qin_Peng_Liu_Li_2020}
Li, Y.; Yao, K.; Qin, L.; Peng, S.; Liu, Y.; and Li, X. 2020{\natexlab{c}}.
\newblock Span-Based Neural Buffer: Towards Efficient and Effective Utilization
  of Long-Distance Context for Neural Sequence Models.
\newblock \emph{Proceedings of the AAAI Conference on Artificial Intelligence}
  34(05): 8277--8284.
\newblock \doi{10.1609/aaai.v34i05.6343}.
\newblock
  \urlprefix\url{https://ojs.aaai.org/index.php/AAAI/article/view/6343}.

\bibitem[{Mairesse and Young(2014)}]{mairesse2014stochastic}
Mairesse, F.; and Young, S. 2014.
\newblock Stochastic language generation in dialogue using factored language
  models.
\newblock \emph{Computational Linguistics} 40(4): 763--799.

\bibitem[{Mikolov et~al.(2010)Mikolov, Karafi{\'a}t, Burget,
  {\v{C}}ernock{\`y}, and Khudanpur}]{mikolov2010recurrent}
Mikolov, T.; Karafi{\'a}t, M.; Burget, L.; {\v{C}}ernock{\`y}, J.; and
  Khudanpur, S. 2010.
\newblock Recurrent neural network based language model.
\newblock In \emph{Eleventh annual conference of the international speech
  communication association}.

\bibitem[{Mirkovic and Cavedon(2011)}]{mirkovic2011dialogue}
Mirkovic, D.; and Cavedon, L. 2011.
\newblock Dialogue management using scripts.
\newblock US Patent 8,041,570.

\bibitem[{Nayak et~al.(2017)Nayak, Hakkani-T{\"u}r, Walker, and
  Heck}]{nayak2017plan}
Nayak, N.; Hakkani-T{\"u}r, D.; Walker, M.~A.; and Heck, L.~P. 2017.
\newblock To Plan or not to Plan? Discourse Planning in Slot-Value Informed
  Sequence to Sequence Models for Language Generation.
\newblock In \emph{INTERSPEECH}, 3339--3343.

\bibitem[{Novikova, Du{\v{s}}ek, and Rieser(2017)}]{novikova-etal-2017-e2e}
Novikova, J.; Du{\v{s}}ek, O.; and Rieser, V. 2017.
\newblock The {E}2{E} Dataset: New Challenges For End-to-End Generation.
\newblock In \emph{Proceedings of the 18th Annual {SIG}dial Meeting on
  Discourse and Dialogue}, 201--206. Saarbr{\"u}cken, Germany: Association for
  Computational Linguistics.
\newblock \doi{10.18653/v1/W17-5525}.
\newblock \urlprefix\url{https://www.aclweb.org/anthology/W17-5525}.

\bibitem[{Oh and Rudnicky(2000)}]{oh2000stochastic}
Oh, A.; and Rudnicky, A. 2000.
\newblock Stochastic language generation for spoken dialogue systems.
\newblock In \emph{ANLP-NAACL 2000 Workshop: Conversational Systems}.

\bibitem[{Ratnaparkhi(2000)}]{ratnaparkhi2000trainable}
Ratnaparkhi, A. 2000.
\newblock Trainable methods for surface natural language generation.
\newblock In \emph{Proceedings of the 1st North American chapter of the
  Association for Computational Linguistics conference}, 194--201. Association
  for Computational Linguistics.

\bibitem[{Sutskever, Vinyals, and Le(2014)}]{sutskever2014sequence}
Sutskever, I.; Vinyals, O.; and Le, Q.~V. 2014.
\newblock Sequence to sequence learning with neural networks.
\newblock \emph{arXiv preprint arXiv:1409.3215} .

\bibitem[{Tran and Nguyen(2017)}]{tran-nguyen-2017-natural}
Tran, V.-K.; and Nguyen, L.-M. 2017.
\newblock Natural Language Generation for Spoken Dialogue System using {RNN}
  Encoder-Decoder Networks.
\newblock In \emph{Proceedings of the 21st Conference on Computational Natural
  Language Learning ({C}o{NLL} 2017)}, 442--451. Vancouver, Canada: Association
  for Computational Linguistics.
\newblock \doi{10.18653/v1/K17-1044}.
\newblock \urlprefix\url{https://www.aclweb.org/anthology/K17-1044}.

\bibitem[{van~den Oord, Vinyals et~al.(2017)}]{van2017neural}
van~den Oord, A.; Vinyals, O.; et~al. 2017.
\newblock Neural discrete representation learning.
\newblock In \emph{Advances in Neural Information Processing Systems},
  6306--6315.

\bibitem[{Vaswani et~al.(2017)Vaswani, Shazeer, Parmar, Uszkoreit, Jones,
  Gomez, Kaiser, and Polosukhin}]{vaswani2017attention}
Vaswani, A.; Shazeer, N.; Parmar, N.; Uszkoreit, J.; Jones, L.; Gomez, A.~N.;
  Kaiser, {\L}.; and Polosukhin, I. 2017.
\newblock Attention is all you need.
\newblock In \emph{Advances in neural information processing systems},
  5998--6008.

\bibitem[{Vinyals, Fortunato, and Jaitly(2015)}]{vinyals2015pointer}
Vinyals, O.; Fortunato, M.; and Jaitly, N. 2015.
\newblock Pointer networks.
\newblock In \emph{Advances in neural information processing systems},
  2692--2700.

\bibitem[{Wen et~al.(2015{\natexlab{a}})Wen, Ga{\v{s}}i{\'c}, Kim,
  Mrk{\v{s}}i{\'c}, Su, Vandyke, and Young}]{wen-etal-2015-stochastic}
Wen, T.-H.; Ga{\v{s}}i{\'c}, M.; Kim, D.; Mrk{\v{s}}i{\'c}, N.; Su, P.-H.;
  Vandyke, D.; and Young, S. 2015{\natexlab{a}}.
\newblock Stochastic Language Generation in Dialogue using Recurrent Neural
  Networks with Convolutional Sentence Reranking.
\newblock In \emph{Proceedings of the 16th Annual Meeting of the Special
  Interest Group on Discourse and Dialogue}, 275--284. Prague, Czech Republic:
  Association for Computational Linguistics.
\newblock \doi{10.18653/v1/W15-4639}.
\newblock \urlprefix\url{https://www.aclweb.org/anthology/W15-4639}.

\bibitem[{Wen et~al.(2015{\natexlab{b}})Wen, Ga{\v{s}}i{\'c}, Mrk{\v{s}}i{\'c},
  Su, Vandyke, and Young}]{wen-etal-2015-semantically}
Wen, T.-H.; Ga{\v{s}}i{\'c}, M.; Mrk{\v{s}}i{\'c}, N.; Su, P.-H.; Vandyke, D.;
  and Young, S. 2015{\natexlab{b}}.
\newblock Semantically Conditioned {LSTM}-based Natural Language Generation for
  Spoken Dialogue Systems.
\newblock In \emph{Proceedings of the 2015 Conference on Empirical Methods in
  Natural Language Processing}, 1711--1721. Lisbon, Portugal: Association for
  Computational Linguistics.
\newblock \doi{10.18653/v1/D15-1199}.
\newblock \urlprefix\url{https://www.aclweb.org/anthology/D15-1199}.

\bibitem[{Zhu, Zeng, and Huang(2019)}]{zhu2019multi}
Zhu, C.; Zeng, M.; and Huang, X. 2019.
\newblock Multi-task Learning for Natural Language Generation in Task-Oriented
  Dialogue.
\newblock In \emph{Proceedings of the 2019 Conference on Empirical Methods in
  Natural Language Processing and the 9th International Joint Conference on
  Natural Language Processing (EMNLP-IJCNLP)}, 1261--1266.

\end{thebibliography}
	\bibliographystyle{aaai21}
	
\end{document}